\documentclass{article}

\usepackage{arxiv}

\usepackage[utf8]{inputenc} % allow utf-8 input
\usepackage[T1]{fontenc}    % use 8-bit T1 fonts
\usepackage{hyperref}       % hyperlinks
\usepackage{url}            % simple URL typesetting
\usepackage{booktabs}       % professional-quality tables
\usepackage{amsfonts}       % blackboard math symbols
\usepackage{nicefrac}       % compact symbols for 1/2, etc.
\usepackage{microtype}      % microtypography
\usepackage{lipsum}		% Can be removed after putting your text content
\usepackage{graphicx}
\usepackage{amsmath}
\usepackage[numbers]{natbib}
\usepackage{doi}

\title{Structured Latent Variable Models for Articulated Object Interaction}

\author{Emily Liu \\
	Massachusetts Institute of Technology\\
	Cambridge, MA 02139\\
	\texttt{emizfliu@mit.edu} \\
	\And
	Michael Noseworthy \\
	Massachusetts Institute of Technology\\
	Cambridge, MA 02139\\
	\texttt{mnosew@mit.edu} \\
	\And
	Nicholas Roy \\
	Massachusetts Institute of Technology\\
	Cambridge, MA 02139\\
	\texttt{nickroy@csail.mit.edu} \\
}

\renewcommand{\shorttitle}{Structured Latent Variable Models for Articulated Object Interaction}

\hypersetup{
pdftitle={Structured Latent Variable Models for Articulated Object Interaction},
}

\begin{document}
\maketitle

\begin{abstract}
	
In this paper, we investigate a scenario in which a robot learns a low-dimensional representation of a door given a video of the door opening or closing. This representation can be used to infer door-related parameters and predict the outcomes of interacting with the door. Current machine learning based approaches in the doors domain are based primarily on labelled datasets. However, the large quantity of available door data suggests the feasibility of a semisupervised approach based on pretraining. To exploit the hierarchical structure of the dataset where each door has multiple associated images, we pretrain with a structured latent variable model known as a neural statistician. The neural satsitician enforces separation between shared context-level variables (common across all images associated with the same door) and instance-level variables (unique to each individual image). We first demonstrate that the neural statistician is able to learn an embedding that enables reconstruction and sampling of realistic door images. Then, we evaluate the correspondence of the learned embeddings to human-interpretable parameters in a series of supervised inference tasks. It was found that a pretrained neural statistician encoder outperformed analogous context-free baselines when predicting door handedness, size, angle location, and configuration from door images. Finally, in a visual bandit door-opening task with a variety of door configuration, we found that neural statistician embeddings achieve lower regret than context-free baselines.
\end{abstract}

\section{Introduction}

\begin{figure}[h]
    \centering
    \includegraphics[scale=0.75]{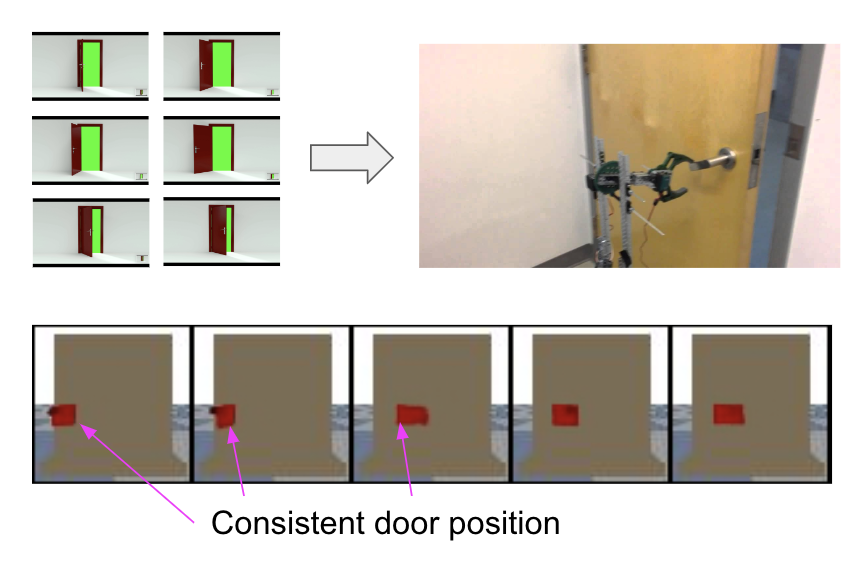}
    \caption{We can take advantage of the hierarchical nature of a door dataset to learn parameters and policies relevant to the door opening task. For example, in the above image it is shown that the same door will have a consistent relative position to the backboard, regardless of its configuration.}
    \label{fig:overview}
\end{figure}

As machine learning algorithms become more robust and effective, it is increasingly popular in robotics to incorporate image data and computer vision strategies to aid in manipulation tasks. One such category of tasks is door manipulation, in which a robot learns how to open a door given visual input. In this class of problems, a CNN-based neural network is typically used to infer parameters from the image (\cite{selforganizing}, \cite{cnnhandles}), which can be paired with a set of action parameters that dictate how the robot interacts physically with the door to predict a reward function (ie, the distance opened) (\cite{versatiledoorop}). However, obtaining these labels can be expensive and infeasible as it requires having all the necessary door information beforehand. Moreover, as doors can vary widely visually (in terms of size, handle location, material, etc), it is difficult to generalize models across doors datasets. However, there is a smaller set of parameters that dictate motion that are more universal across doors. These two observations suggest the use of a semi-supervised learning scheme wherein we pretrain a latent variable model on a large unlabelled dataset in order to learn low-dimensional representations of the door images, and then finetune to specific parameter and reward inference tasks after a low-dimensional representation is learned.

Representing complex visual data in lower dimensions by use of a latent variable model allows for more efficient task learning and enables generalization to unseen instances of the task. However, one current issue with semisupervised models is that it is difficult to determine the exact dependence relationship between the learned variables. This in turn makes it difficult to interpret these variables, leading to less efficient finetuning. Moreover, a single image may not necessarily fully encode all relevant parameters of a door. We need a way to consistently predict parameters shared across images.

To address this problem, we note that the variables in our task exhibit a structured relationship. In downstream tasks, we often only care about interacting with a single object at a time, meaning that many parameters corresponding to the door itself (for example: size, width, etc) stay constant throughout the task. In our case, only the door configuration changes throughout the task. However, if the learned variables do not reflect this dependence relation, it may require more adaptation data in the downstream task than if the conditional relationship is reflected. 

We can take advantage of the hierarchical structure of the doors dataset to enforce separation between object-level and configuration-level variables (Fig \ref{fig:overview}). Most robotic cameras have access to a stream of video data, meaning that it is easy to obtain multiple images of the same object in different configurations: in the doors domain, we would simply sample near-consecutive frames from a video of an opening door. As a result, we have a structured dataset where each door has multiple associated images. In other words, the dataset has two levels of parameters: object-level parameters, which remain constant across multiple images of the same object (the door itself), and configuration-level parameters, which change according to the object's configuration (such as the angle to which the door is open).

To learn a parameterization of this hierarchical dataset, we propose to pretrain using the Neural Statistician (NS) latent variable model \cite{ns}. The neural statistician learns two sets of latent variables from a collection of images: context-level variables, which are shared across all images in the collection, and instance-level variables, which are conditioned on context-level parameters and are different for each image. In our dataset, each collection consists of images of the same door in different configurations. The learned context-level parameters should correspond to the object itself, and the learned instance-level parameters should correspond to the object configuration.

It is known that latent variable models can be used as generative models by decoding values sampled from the latent variable space. We demonstrate that using the Neural Statistician model to generate samples enables us to keep object-level details constant by only sampling at the instance-level. 

In addition to learning latent variable representations, we investigate how well a pretrained neural statistician lends itself to downstream manipulation and parameter prediction tasks, facilitating exploration of new objects. Learning a latent variable representation ideally allows for finetuning on fewer samples. To better understand the advantages provided by the neural statistician, we compare the representations learned by the neural statistician with those learned by a context-free CNN baseline and a variational autoencoder. The methods used in this project can be extended to more complex domains where known effective parametrizations are less well established.

\section{Related Work}

% TODO:
% \begin{enumerate}
%     \item DoorGym paper - RL
%     \item CPP paper - online learning (100 doors, 100 interactions)
%     \item Note that our method is an alternative to online/RL and requires less structured cnn at expense of more pretraining data
% \end{enumerate}

Multiple CNN-based approaches to detect, locate, and manipulate door handles have been explored in the past. In \cite{selforganizing}, a multi-step system of edge detection, line detection, feature extraction, and door detection was implemented with the help of preexisting labels. Door axes were configured to be vertical to the parallel axis. In \cite{cnnhandles}, a convolutional neural network was used for door ROI detection and the handle was located using a point cloud derived from the CNN learned features.  Similarly, in \cite{versatiledoorop}, a YOLONet was used to detect the door and handle, the ROI of which was then used in a point cloud based handle maniupation task.

The majority of work for detection and interaction with doors relies on labelled datasets and supervised learning. Few attempts have been made to apply semisupervised learning techniques to the doors domain. However, there is a large quantity of unlabelled door data that can be used in a semisupervised context to pretrain models to better perform on downstream door detection/opening tasks. We will investigate the use of latent variable models in the doors domain. Since door images have a hierarchical structure of parameters, we believe that using a hierarchical latent variable model will allow for further improvement in the domain.

Pretrained latent variable models additionally have the potential to aid in neural network based policy learning tasks by reducing the number of interactions required. In \cite{cpp}, random interactions with a door required 100 interactions per 100 doors to successfully open a door when using a CNN with 2D feature points. We aim to demonstrate in this work that pretraining with the hierarchical model achieves success using fewer objects in the training set, using a less structured finetuning network.

% IN PROGRESS: I found that the method in \cite{cpp} leads to much lower regret than our method so i'm figuring out how to debug our setup. Will come back to this after that

\section{Methods}

\subsection{Domain}

\begin{figure}[h]
    \centering
    \includegraphics[scale=0.4]{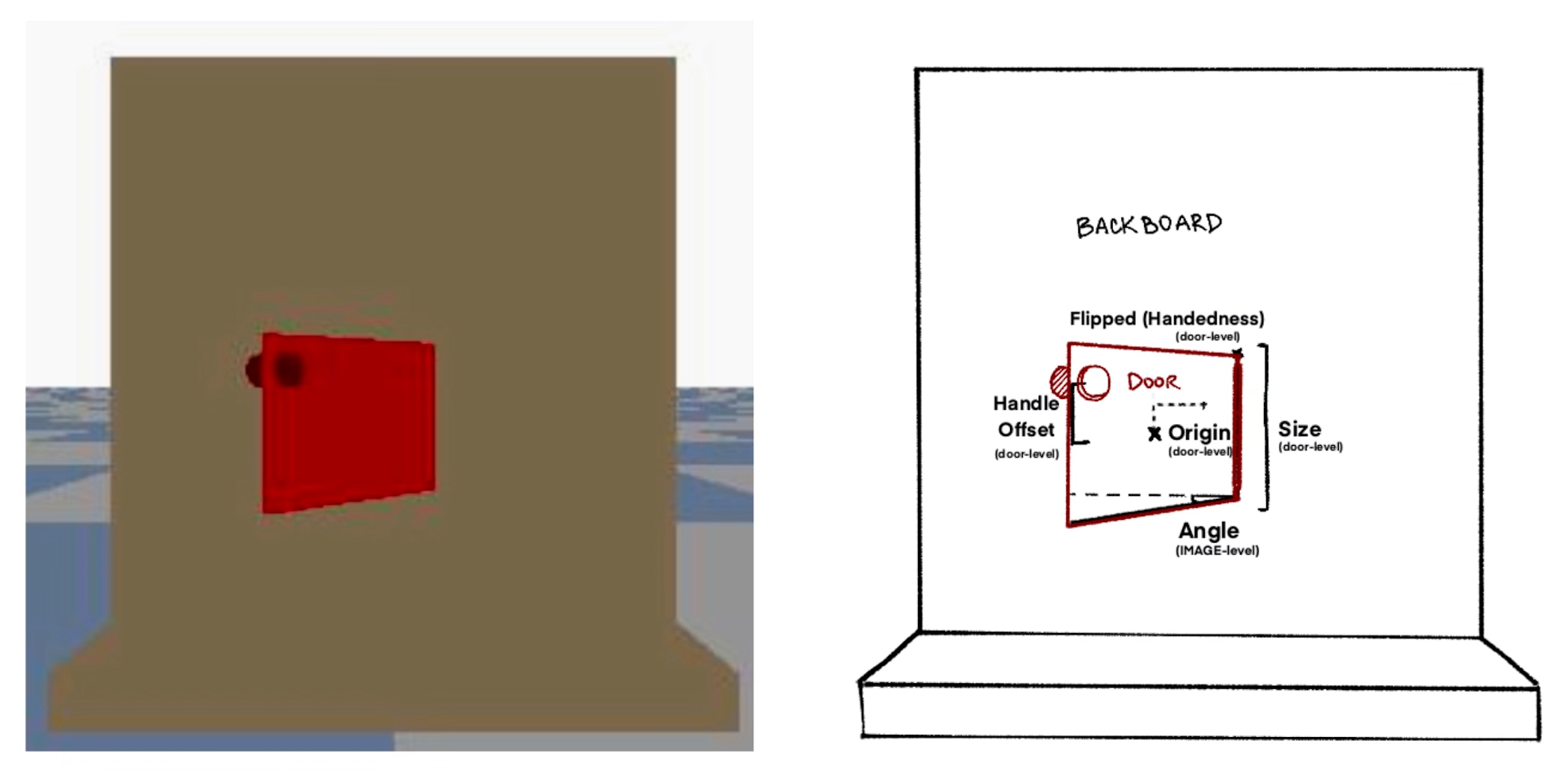}
    \caption{Example of original input image (a), line drawing with labelled learnable parameters (b).}
    \label{fig:door_params_labelled}
\end{figure}

We consider a domain of cabinet-like doors in which the door itself comprises a small portion of the image and is always fully visible against a backboard (Figure \ref{fig:door_params_labelled} left). We assume that the camera operates from a fixed viewpoint, so we see the backboard and all doors head-on. \footnote{
In practice, we assume that the robot can localize itself with respect to the backboard such that the viewpoint resembles that of our dataset. We additionally assume that the video from which the agent samples visual input has minimal occluding objects (hands, etc) that can be removed from the image via image inpainting/object removal techniques \cite{imagecompletion}, so the only relevant object in the image is the door itself.} For the sake of simplicity, we work with a simplified domain in which the door is consistently red and the backboard brown as seen in Figure \ref{fig:door_params_labelled}a. However, once proven useful, the methods from this work can be applied to more general datasets where the door and backboard are not guaranteed to be any particular color, and to other doors datasets as well, such as datasets of doors in rooms where the door may occupy a larger portion of the image and the backboard is less well defined. These datasets are not considered in this work but are potential areas of further study.

Our data space consists of doors, which are parameterized by their size, location, handle offset, and if they open to the left or the right (handedness). From each door, we take a fixed number of camera images of the door opened at different angles. See the line drawing in Figure \ref{fig:door_params_labelled}b for a visualization of all door parameters. Images are grouped by the door from which they are sampled and then passed through the neural statistician model.

We used the physics engine PyBullet (\cite{pybullet}) to simulate the door environment and generate our datasets. Doors were generated on a backboard and opened to a random angle between 0 and 180 degrees. Each door was sampled five times, making for five images per door.

\subsection{Task}

Our task consists of two phases: the pretraining phase and the finetuning phase. In the pretraining phase, we train a latent variable model on a large unlabelled doors dataset to obtain low-dimensional hierarchical embeddings. We then assess the quality of these embeddings via reconstruction and sampling tasks. In the finetuning phase, we use the pretrained encoder weights and architecture to learn two supervised tasks on small labelled datasets. The first supervised task is a parameter inference task wherein the model attempts to predict door parameters based on the learned embeddings. This task serves to provide more insight towards what information is learned in the latent space and to what level parameters are disentangled. If a given door parameter is well-encoded in the latent space, we would expect the model to be able to infer it with low error during finetuning. In the second supervised task, a robotic arm is allowed to interact with a closed door and we use a model to predict the distance opened by the door based on the action parameters and encoded visual input from the latent variable model, again using the pretrained encoder to process the visual input from the finetuning task.

\subsection{Models}

\subsubsection*{Variational Autoencoder}

\begin{figure}[h]
    \centering
    \includegraphics[scale=0.45]{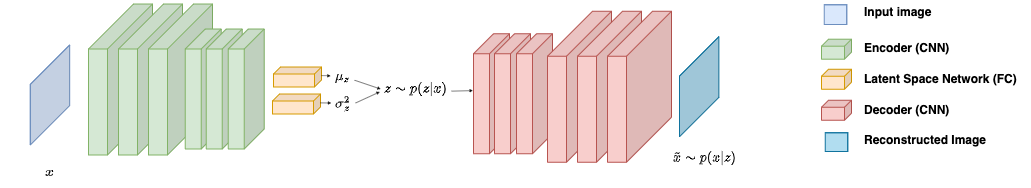}
    \caption{Variational Autoencoder Network Architecture.}
    \label{fig:vae_arch}
\end{figure}

The hierarchical Neural Statistician model is based on the Variational Autoencoder model \cite{vae}. For observed $\mathbf{x}$, the VAE learns a latent variable $\mathbf{z}$ according to the probability distribution
\begin{align*}
    p(\mathbf{x}) = \int p(\mathbf{x} | \mathbf{z}; \theta) p(\mathbf{z}) d\mathbf{z}.
\end{align*}

The parameters $\theta$ are learned through a neural network encoder $q(\mathbf{z} | \mathbf{x}; \phi)$ and the model is optimized using the evidence lower bound as a loss function:
\begin{align*}
    \mathcal{L}_x = \mathbf{E}_{q(\mathbf{z} | \mathbf{x}; \phi)}\left[\log p(\mathbf{x} | \mathbf{z}; \theta)\right] - D_{KL}(q(\mathbf{z} | \mathbf{x}; \phi) || p(\mathbf{z}))
\end{align*}

\subsubsection*{Neural Statistician}

\begin{figure}[h]
    \centering
    \includegraphics[scale=0.35]{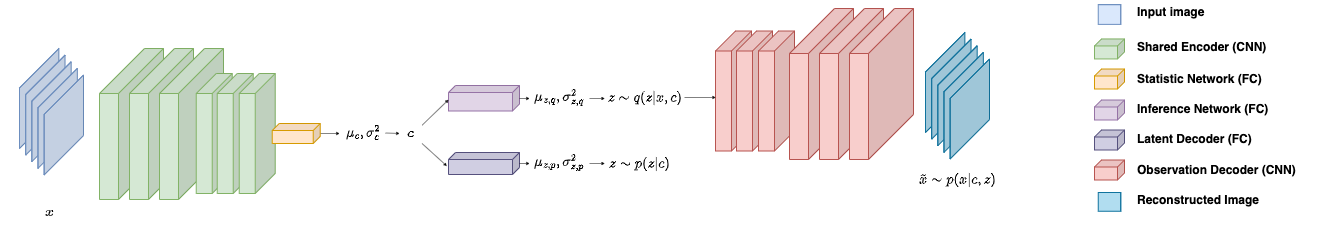}
    \caption{Neural Statistician Network Architecture. Note how there are multiple layers of sampling in the latent space, corresponding to the multiple latent variables.}
    \label{fig:ns_arch}
\end{figure}

While the variational autoencoder model provides a good way of learning a low-dimensional representation of a complex dataset, it has several shortcomings. First, it is unable to learn connections between multiple images, meaning there is no way for the VAE model to know that a collection of images all belong to the same door. Second, the features learned by the VAE are not guaranteed to be disentangled, so it is difficult to separate object-level and configuration-level parameters.

We used a neural statistician (NS) model \cite{ns} to learn the door embeddings. The neural statistician model extends the variational autoencoder model to enable learning on datasets (as opposed to individual data points). Elements within one dataset are grouped together by a context variable ($c$), which captures the parameters associated with the likelihood of the dataset itself. Each individual data point is then parameterized by an instance level latent variable $z$. The likelihood for one dataset $D$ is given by
\begin{align*}
    p(D; \theta) = \int p(\mathbf{c}) \left[\prod_{\mathbf{x} \in D} p(\mathbf{x} | \mathbf{z}; \theta)p(\mathbf{z} | \mathbf{c}; \theta) d\mathbf{z}\right] d\mathbf{c}
\end{align*}

where $\theta$ parameterizes the encoder and decoder, $p(\mathbf{z} | \mathbf{c}; \theta)$ is a Gaussian with diagonal covariance learned via a neural network, and $p(\mathbf{x} | \mathbf{z}; \theta)$ is learned through another observation network \cite{ns}. 

The neural statistician is trained using the variational lower bound on log likelihood as an objective function. This is given by
\begin{align*}
    \mathcal{L}_D = \mathbf{E}_{q(\mathbf{c} | D, \phi)} \left[ \sum_{\mathbf{x} \in D}\mathbf{E}_{q(\mathbf{z}|\mathbf{c}, \mathbf{x}; \phi)}[\log p(\mathbf{x} | \mathbf{z}; \theta)] - D_{KL}(q(\mathbf{z}|\mathbf{c}, \mathbf{x}; \phi) || p(\mathbf{z} | \mathbf{c}; \theta))\right] - D_{KL}(q(\mathbf{c}|D; \phi) || p(\mathbf{c})))
\end{align*}
where $q(\mathbf{z}|\mathbf{c}, \mathbf{x}; \phi)$ and $q(\mathbf{c}|D; \phi)$ are diagonal Gaussians learned by neural networks. \cite{ns}

The neural statistician architecture consists of five parts:
\begin{enumerate}
    \item A shared \textbf{encoder} network ($\mathbf{x} \rightarrow \mathbf{h}$) consisting of convolutional layers that accepts the original collection of images as input and outputs shared embeddings across all input images.
    \item A fully-connected \textbf{statistic} network ($q(\mathbf{c} | D, \phi)$; $\mathbf{h} \rightarrow \mu_\mathbf{c}, \sigma_\mathbf{c}$) that predicts context variables based on $h$ from the encoder network.
    \item A fully-connected \textbf{inference} network that infers the latent variable $\mathbf{z}$ ($q(\mathbf{z} | \mathbf{x}, \mathbf{c}, \phi)$; $\mathbf{h}, \mathbf{c} \rightarrow \mu_\mathbf{z}, \sigma_\mathbf{z}$) conditioned on context $c$.
    \item A fully-connected \textbf{latent decoder} network ($p(\mathbf{z} | \mathbf{c}, \theta)$; $\mathbf{c} \rightarrow \mu_\mathbf{z}, \sigma_\mathbf{z}$) that samples values of $z$ from a given context $c$.
    \item A convolutional \textbf{observation decoder} network ($p(\mathbf{x} | \mathbf{c}, \mathbf{z}, \theta)$; $\mathbf{c}, \mathbf{z} \rightarrow \mu_\mathbf{x}$) that generates reconstructed images from latent $z$ and $c$ embeddings from the statistic and inference networks.
\end{enumerate}

The statistic, inference, and latent decoder networks all consist of fully connected linear layers with 1,000 units and ELU activations and a final fully connected layer to mean and variance predictions. A visualization of the NS architecture can be found in Figure \ref{fig:ns_arch}.

\section{Experiments}
\subsection{Results summary}
In this section, we observe the following results:
\begin{enumerate}
    \item The neural statistician effectively learns the structured latent space in the doors dataset and conditional sampling of the $z$ latent variable with constant $c$ allows for generation of images of the same door (shared object-level parameters) with different angles (instance-level parameters).
    \item A finetuned neural statistician encoder is better able to infer human-interpretable hierarchical door parameters than pretraining on a non-structured VAE baseline due to learning shared object-level features across multiple images of the same door.
    \item Latent embeddings learned by the neural statistician can be used in a policy selection task where the model is presented with a series of door images and selects the action out of a set of possible door-opening policy that yields the lowest predicted regret. When the neural statistician is used, the model was able to select actions that achieve lower regret (and better predict the reward function) when compared to baselines, especially when the door images are of the same door open to different angles.
\end{enumerate}

\subsection{Baselines}

We compare the performance of the neural statistician to a variational autoencoder (VAE) baseline, trained on the same dataset, wherein the dimension of the latent space matches the total dimension of the NS $\mathbf{c}$ and $\mathbf{z}$ variables combined. The VAE encoder has the same architecture as the shared encoder in the NS and the decoder consists of a fully connected layer followed by the architecture from the NS observation decoder (Figure \ref{fig:vae_arch}). Additionally, we compare the finetuned performance of both NS and VAE to a vanilla CNN baseline with no pretraining, using the same architecture as the convolutional encoders of the NS and VAE.
\subsection{Latent Variable Embeddings}
\label{sec:model_eval}

Before we address parameter inference and reward prediction tasks, we need to pretrain the latent variable models. The neural statistician and variational autoencoder were pretrained on differently sized training sets of 400, 800, 4000, and 8000 doors. Each door has  associated images, each with a different configuration (angle). The quality of these embeddings were evaluated first through image reconstruction, and then through generative sampling from the latent space. All images used are of size 64$\times$64. In order to demonstrate efficacy with the minimal number of latent dimensions and to better interpret the learned variables, we used low bottleneck sizes for the models. We set parameters $c = 8, z = 1$ in the neural statistician to match the 6 object-level and 1 configuration-level parameter used to generate each door image (by convention, we rounded 6 up to the nearest power of 2).

\subsubsection*{Image Reconstruction}
We first assess the performance of the latent variable models in an image reconstruction task on unseen doors. Since reconstruction loss is used to train the network, achieving high quality reconstructions is an indication that the model was able to learn a low-dimensional representation of the input distribution. Moreover, by examining where the reconstruction differs from the original, we can gain insight as to where the model may perform well or struggle during finetuning.

\begin{figure}[h]
    \centering
    \includegraphics[trim={0 .3cm 0 0},clip, scale=0.9]{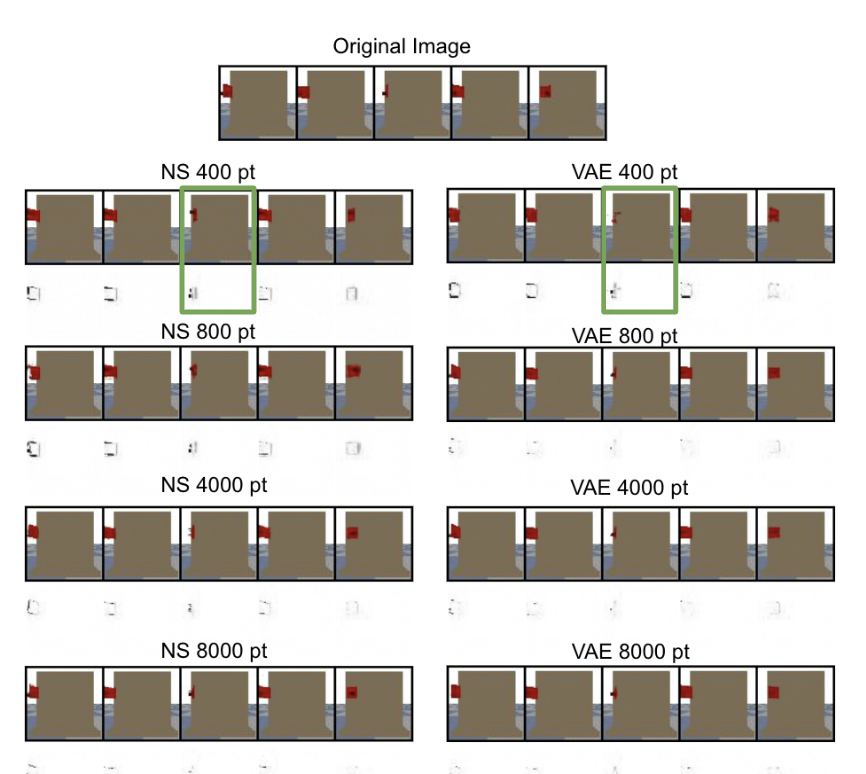}
    \caption{Reconstructions from NS and VAE trained on 400, 800, 4000, and 8000 doors. Pixel diffs (below each set of images) are evaluated as the absolute difference between the original image and reconstruction. Note the difference in the third reconstructed image from the NS and VAE models pretrained on 400 doors: while the VAE struggles to capture the shape of the door due to lack of context, the NS learns the shape of the door and the error results from the slightly shifted door configuration (angle).}
    \label{fig:reconstructions}
\end{figure}

Figure \ref{fig:reconstructions} shows us that the quality of the reconstructions improves with the size of the dataset used to train the model, as expected. The biggest discrepancy between the performance of the neural statistician and the VAE is on the models pretrained with 400 doors, evaluating the door open at 90 degrees (the third door in the original set of images). The variational autoencoder has difficulty reconstructing this door because it is unable to infer context from the other images, meaning that a lot more images are required for pretraining for the network to capture this nuance. On the other hand, the neural statistician is able to get the shape of the door even with a low number of pretraining samples, the error mostly coming from the angle of the door being different than the angle in the original image.

\subsubsection*{Sampling}

\begin{figure}[h]
    \centering
    \includegraphics[scale=0.4]{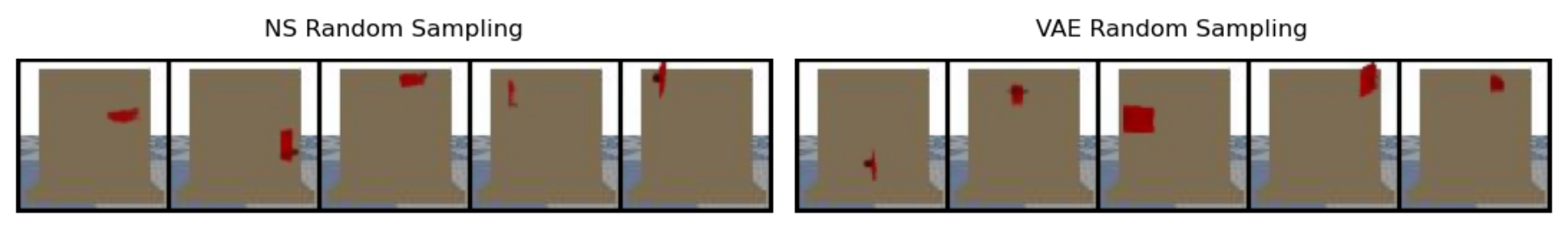}
    \caption{Randomized sampling of latent space from NS (left) and VAE (right), both trained on 8000 doors.}
    \label{fig:random_sample}
\end{figure}

After we confirm that both neural statistician and variational autoencoder models can reconstruct images without significant loss, we evaluate the quality of the latent space itself by decoding random normal samples from the low-dimensional representation (where the mean and standard deviation are inferred from those of the original dataset encodings). As can be seen in \ref{fig:random_sample}, both the neural statistician and variational autoencoder can produce realistic-looking doors from random samples from the latent space, meaning that there is good utilization of the latent space in both models.

\begin{figure}[h]
    \centering
    \includegraphics[scale=0.6]{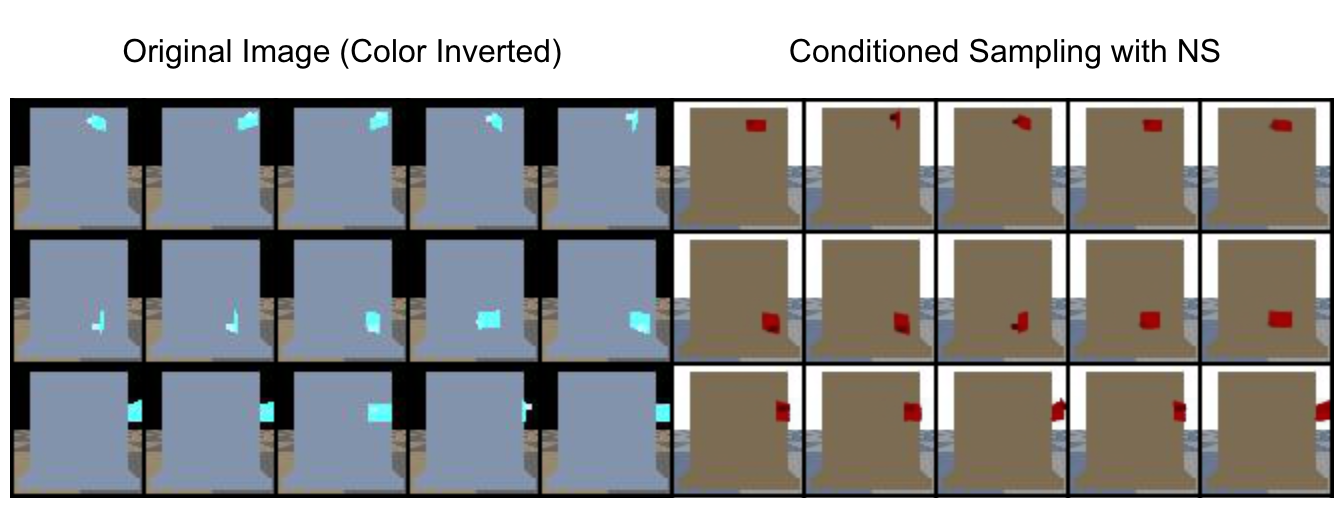}
    \caption{Conditioned sampling of the $z$ parameter from neural statistician trained on 8000 doors. Left: Original door images, inverted color; Right: samples generated using the same $c$ as the left image set on the same row.}
    \label{fig:conditioned_sample}
\end{figure}

However, we can further demonstrate for the neural statistician that the hierarchical structure of the latent space allows us to more easily control the scope of our sampling. By holding context parameters constant, we can simulate taking multiple snapshots of a single door. Figure \ref{fig:conditioned_sample} shows images generated by the NS via random sampling of the $z$ parameter with the trained model using the context-level $c$ parameters from a given set of door snapshots. Sampling the $z$ parameter manifests as a visual change in door angle, indicating that the neural statistician was able to learn common latent parameters across the five images from each door.

\begin{figure}[h]
    \centering
    \includegraphics[scale=0.5]{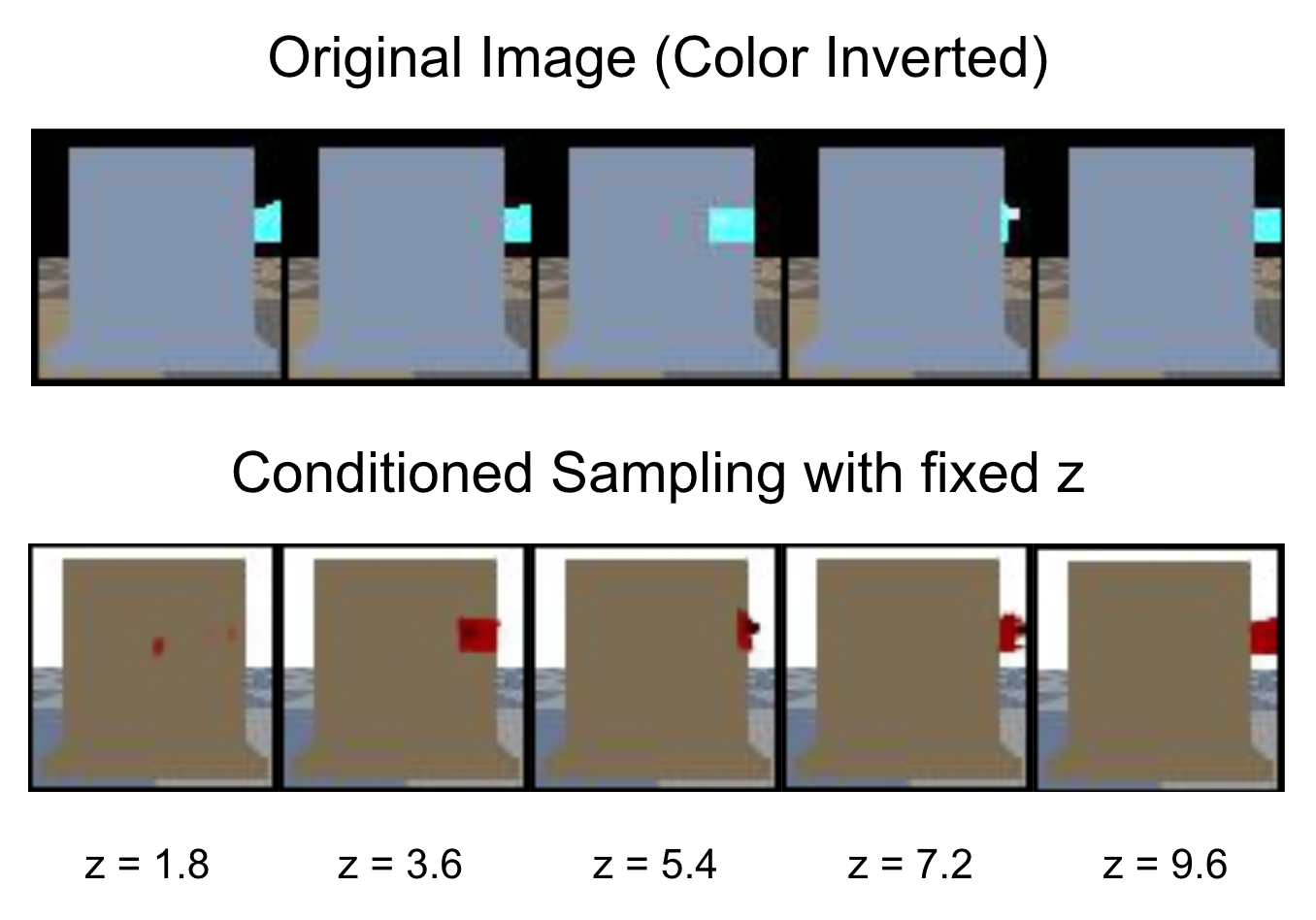}
    \caption{Sampling from latent space, letting z = [-10, 9.6] in intervals of 0.4. Left: the input sample (repeated for padding), Right: images generated from conditional sampling.}
    \label{fig:ns_var_z}
\end{figure}

Ideally, the instance-level parameter $z$ should correspond with the angle of the door. To evaluate this, we fix $c$ and sample from $z$. Figure \ref{fig:ns_var_z} shows samples generated from the NS trained on 8000 doors letting $z$ vary in the range [1.8, 9.6] in intervals of 1.8. It is observed that small changes in $z$ yield gradual changes in door angle along with door location. This correlation will prove useful for finetuning.

\subsection{Parameter Inference}

\begin{figure}[h]
    \centering
    \includegraphics[scale=0.5]{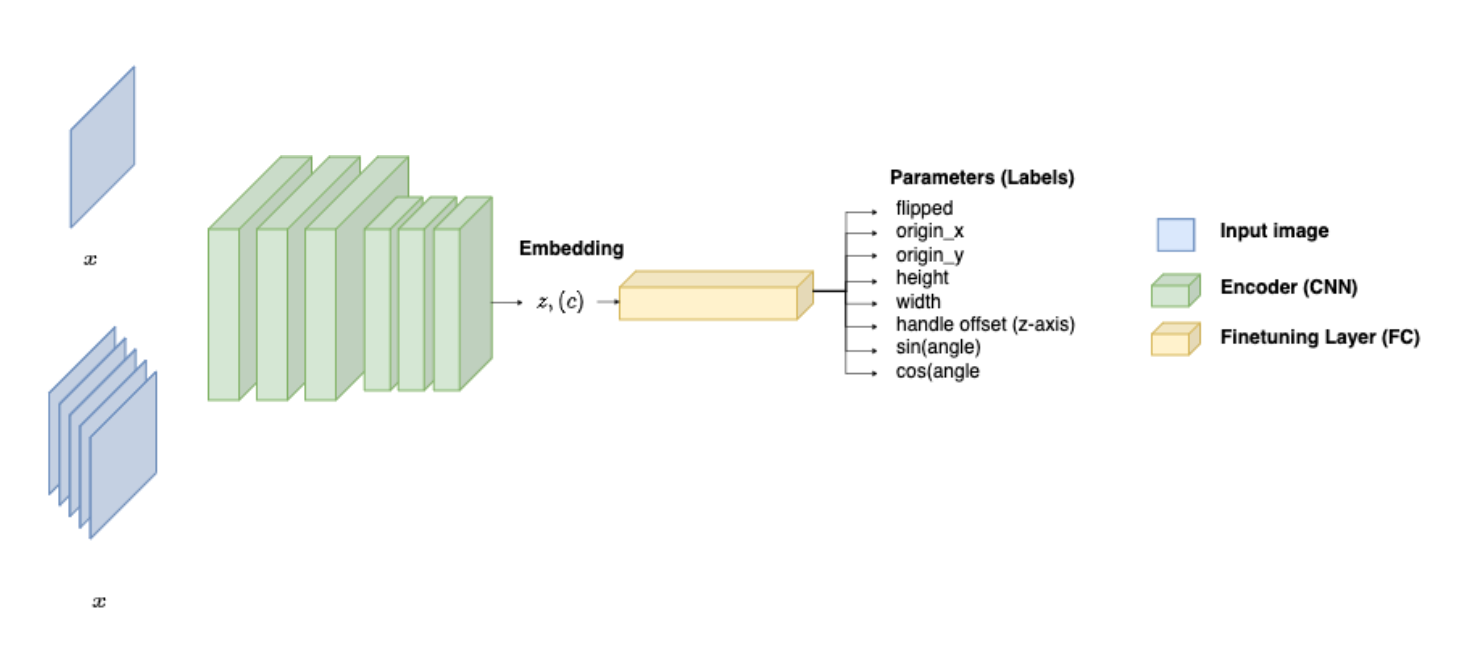}
    \caption{Finetuning architecture for parameter inference task. Note how images are passed individuallly under the VAE setting and as a group under the NS setting.}
    \label{fig:ft_params}
\end{figure}

Our first finetuning task is to predict the parameters that were used to generate the door images (angle (in terms of sine and cosine), direction of opening (handedness), origin, size, and handle offset, as indicated in Figure \ref{fig:door_params_labelled}). Door parameters are predicted from camera images when training on the learned latent variables from the neural statistician or the variational autoencoder. While parameter inference is not a standardized task in existing literature since there is no universally recognized way to parameterize a door, a latent variable model is able to perform better on parameter inference tasks if it can infer object-level context cues from an image. Therefore, we expect the pretrained neural statistician to outperform the VAE as it can infer context, while the VAE cannot.

For both the NS and VAE, we finetuned to a supervised training set of 400 doors with five samples each. For the neural statistician, the embeddings $c$ and $z$ are computed from the finetuning dataset and then passed through a fully connected layer to predict the door parameters (Figure \ref{fig:ft_params}). We additionally trained on a separate vanilla CNN task where the NS or VAE encoder architecture was kept the same and the weights reinitialized. This was to observe any performance benefits solely from context-dependent set based learning as opposed to from the latent variable representations. We evaluated on a test set of 200 unseen doors and observed the error in prediction for each door parameter.

\begin{table}
    \centering
    \begin{tabular}{c c|c c c}
        \textbf{Model} & \textbf{n\_pretrain} & \textbf{Flipped Accuracy} & \textbf{Origin RMSE} & \textbf{Size RMSE} \\
        \hline
        \hline
        Baseline & --- & $0.713 \pm 0.018$ & $0.029 \pm 0.002$ & $0.018 \pm 0.000$ \\
        \hline
        NS & 400 & $0.638 \pm 0.039$& $0.041 \pm 0.002$ & $0.017 \pm 0.001$\\
        VAE & 400 & $0.655 \pm 0.014$& $0.035 \pm 0.001$ & $0.018 \pm 0.000$\\
        \hline
        NS & 800 & $0.817 \pm 0.034$& $0.043 \pm 0.003$ & $0.016 \pm 0.001$\\
        VAE & 800 & $0.693 \pm 0.013$& $0.032 \pm 0.002$ & $0.017 \pm 0.000$\\
        \hline
        NS & 4000 & \textbf{0.929 $\pm$ 0.010}& $0.036 \pm 0.003$ & $0.015 \pm 0.001$\\
        VAE & 4000 & $0.739 \pm 0.009$& \textbf{0.028 $\pm$ 0.001} & $0.016 \pm 0.000$\\
        \hline
        NS & 8000 & $0.910 \pm 0.011$& $0.038 \pm 0.003$ & \textbf{0.014$\pm$ 0.000}\\
        VAE & 8000 & $0.689 \pm 0.020$& $0.033 \pm 0.001$ & $0.016 \pm 0.000$\\
        \hline
        \\
        \hline
        \textbf{Model} & \textbf{n\_pretrain} & \textbf{Handle Offset RMSE} & \textbf{Angle Degree Error}\\
        \hline
        \hline
        Baseline & --- &  $0.016 \pm 0.001$ & $14.84 \pm 0.71$\\
        \hline
        NS & 400 & $0.026 \pm 0.001$ & $19.78 \pm 2.06$\\
        VAE & 400 & $0.018 \pm 0.002$ & $15.78 \pm 1.01$\\
        \hline
        NS & 800 & $0.017 \pm 0.000$ & $16.43 \pm 0.92$\\
        VAE & 800 & $0.015 \pm 0.001$ & $15.22 \pm 0.77$\\
        \hline
        NS & 4000 & $0.014 \pm 0.001$ & $11.47 \pm 1.23$\\
        VAE & 4000 & $0.016 \pm 0.001$ & $15.25 \pm 0.69$\\
        \hline
        NS & 8000 & \textbf{0.012 $\pm$ 0.001} & \textbf{8.38 $\pm$ 1.06}\\
        VAE & 8000 & $0.017 \pm 0.001$ & $16.06 \pm 0.83$\\
        \hline
    \end{tabular}
    \caption{All finetuning results (mean and standard deviation). Note that the NS exceeds VAE performance for all human-interpretable parameters except for origin, which is not truly context-level in the same way as the other parameters since it depends on both context and instance level features (the apparent origin depends on the true origin and the angle the door is open to), meaning the separation provided by the NS is of less help.}
    \label{tab:all_results}
\end{table}

Table \ref{tab:all_results} shows the result of finetuning the different neural statistician and variational autoencoder models pretrained on 400, 800, 4000 and 8000 doors, along with the baseline performance of a CNN trained directly on the supervised task.

We observe that the average accuracy of the handedness prediction (flipped) is far higher in the neural statistician than the VAE or the baseline. Additionally, the RMSE for the door size is significantly smaller when using the pretrained NS encoder than when using the pretrained VAE encoder. This reflects the initial hypothesis that the neural statistician observes benefit by inferring context from batches of related images.

When we are trying to determine the handedness of a door, without context the only clue we have is the placement of the door's handle (whether it is on the left or right side of the door). When the angle of the door changes, the relative displacement between the door handle and axis also changes, which makes it handedness a difficult parameter to learn on its own. However, when we have multiple images of the same door open at different angles, we know that the door axis must remain the same across all images and therefore the changing of the handle location with the angle can be used to determine handedness.

As expected, pretraining on 8000 doors yields an improvement over the baseline almost unilaterally, but pretraining on smaller samples (400 doors, 800 doors) leads to higher RMSE in the handle offset task and higher MAE in the angle inference task when compared to the VAE and to baseline. The reason for this could be that using the neural statistician to pretrain on too few doors causes convergence at local minima where the separation of c and z parameters does not clearly correspond to any of the defined door parameters.

However, the neural statistician outperforms the CNN baseline on these metrics when the pretraining set is sufficiently large. In angle prediction, this result is in line with human-interpretable parameters. Of all parameters, angle is the parameter that was most associated with the $z$ variable in the neural statistician, as noted in Section \ref{sec:model_eval}. The fact that the neural statistician outperforms the baseline on angle prediction indicates that the forced parameter hierarchy of the NS is useful for separating image-level parameters from context-level parameters.

The handle offset is a similar story. We want to determine how far the handle is (along the z-axis) from the x-y plane that bisects the door. When given a single image, the apparent handle offset is a function that is dependent on angle, as a door opened close to 90 degrees will experience a greater perspective distortion with respect to the camera, meaning the apparent handle offset will be larger than the true handle offset. On the other hand, if we pass a set of related images in a collection, the apparent handle offset in each image will vary slightly as a function of angle, which means that the neural statistician is better able to infer handle offset as it is a function of angle. Similarly, predicting size from a single image int he CNN can be tricky if the angle is close to 90 degrees, meaning that information about the door width is obscured from the camera. The neural statistician has access to a wider range of angles, meaning it is able to make more accurate size calculations.

The one parameter prediction task in which the NS performs worse than the baseline is origin prediction. In this case, learning the coordinates of the door center may be tricky for the neural statistician as the effective door center changes every image, meaning it wasn't truly object-level in the sense that it would be invariant across a dataset.

\subsection{Policy Selection}

\begin{figure}[h]
    \centering
    \includegraphics[scale=0.5]{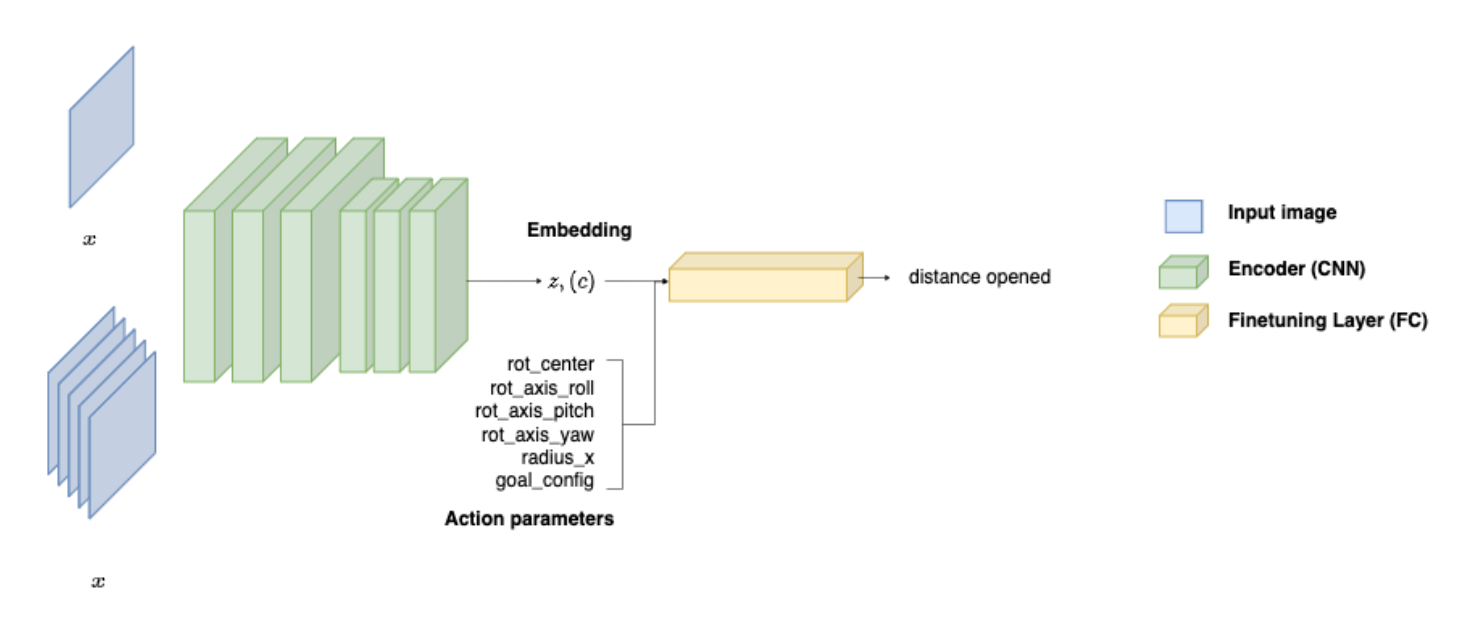}
    \caption{Finetuning architecture for policy learning task. Input is single-image under the VAE setting and grouped images under the NS setting.}
    \label{fig:ft_actions}
\end{figure}
\subsubsection{Problem setup}
Finally, we assess the ability of the pretrained models to aid in learning a door-opening task by predicting the expected regret for a set of door-opening actions, in a similar task to \cite{cpp}. In addition to including the embeddings in finetuning, we pair each input with a set of action parameters to pass into the finetuning layers (Figure \ref{fig:ft_actions}). This task is most similar to what an end application of the neural statistician in robotic manipulation would entail. 

Given a finetuning dataset of 80 doors, we randomly generate 100 different sets of action parameters corresponding to 100 different door-opening policies. One set of parameters consists of the planned (vertical) rotation axis, the planned rotation radius, and the planned goal configuration and opens the door from a closed state. This is equivalent to allowing the robot to interact with 80 different doors 100 times each by randomly calculating and executing 100 different door-opening trajectories. We are able to calculate the actual distance opened by the door for each of these trajectories to generate labels (rewards) for the (image, action) pair passed into the networks.

It is assumed the robot knows the location of the handle of the door and can manipulate it directly. While this reward prediction task is independent of the previous parameter inference task, we find that the advantages provided by the neural statistician in parameter prediction carry over to reward prediction and policy selection as well.

We use two training datasets for this task. The first dataset consists of images of closed doors, as this would be most similar to the visual input from the camera at the start of the door-opening task. The second dataset is more generalized and the doors can be open at any angle. This is more similar to the task initially described, where a robot can learn how to open a door given a video of that door being opened.

To evaluate, we first examine the reward prediction accuracy from the neural network using a test set of 20 doors, also with 100 interactions each. Then, we compute the normalized regret (using the same test set) from the predicted optimal-reward policy, defining the oracle as the maximum distance opened by the door from the 100 trials:
\begin{align*}
    R(a) = \frac{r^* - r(\text{NNET}(X, a))}{r^*}
\end{align*}
where  $r(a)$ is the reward function, $r^*$ is the maximum distance opened out of the sampled actions (such that the minimum possible regret is, $X$ is all the visual input for a door, and NNET($X, a$) is the reward prediction from our model. We first naively evaluate the regret from the best action as predicted by the network ($R_{\text{top}}$).

Because the top prediction may not be robust, we additionally looked at an average of the top 5 actions ($R_{\text{top}5}$). This is equivalent to the expected regret if the agent were to randomly select one of 5 highest ranked actions. 

Based on previous observations, we note that using more samples during pretraining leads to more significant improvements during finetuning. Therefore, for the reward prediction task we finetune solely on the NS and VAE models trained on 8000 doors, using the same context-free CNN baseline trained from scratch.

\begin{table}
    \centering
    \begin{tabular}{c c|c c c}
        \textbf{Input} & \textbf{Model} & \textbf{Reward RMSE} &  \textbf{$R_{\text{top}}$} & \textbf{$R_{\text{top}5}$} \\
        \hline
        \hline
        Closed-door & Baseline &  0.028 $\pm$ 0.000 & 0.697 $\pm$ 0.000 & 0.679 $\pm$ 0.000\\
        Closed-door & VAE & \textbf{0.025 $\pm$ 0.00}3 & 0.492 $\pm$ 0.153 & 0.475 $\pm$ 0.147\\
        Closed-door & NS & 0.026 $\pm$ 0.003 & \textbf{0.414 $\pm$ 0.147} & \textbf{0.435 $\pm$ 0.124}\\
        \hline
        Open-door & Baseline & 0.025 $\pm$ 0.000 & 0.907 $\pm$ 0.000 & 0.682 $\pm$ 0.000\\
        Open-door & VAE & 0.024 $\pm$ 0.001 & 0.721 $\pm$ 0.264 & 0.576 $\pm$ 0.150\\
        Open-door & NS & \textbf{0.022 $\pm$ 0.002} & \textbf{0.472 $\pm$ 0.188} & \textbf{0.382 $\pm$ 0.123}\\
        \hline
    \end{tabular}
    \caption{Reward error and regret on closed-door and open-door input. On closed-door reward prediction, the VAE exceeds NS slightly, but on all other tasks, the NS is able to achieve better results. On both open and closed door tasks, the NS can select for lower regret actions than baselines.}
    \label{tab:eval_policy}
\end{table}

% & 0.522 $\pm$ 0.000
% & 0.284 $\pm$ 0.173
% & 0.285 $\pm$ 0.184
% & 0.429 $\pm$ 0.000
% & 0.354 $\pm$ 0.106
% & 0.186 $\pm$ 0.103

\subsubsection{Results}
Table \ref{tab:eval_policy} lists the RMSE, in meters, for reward prediction from the network and the regret evaluations for the NS and VAE pretrained on 8000 doors along with the CNN baseline for both closed-door and open-door finetuning. We look at the regret of the predicted top action from each model, as well as the average regret of the top 5 actions. We observe that the NS achieves significantly lower regret than baseline and VAE in both the closed-door and open-door tasks, indicating that the structural pretraining was useful for transfer to the manipulation task which also takes advantage of the door structure. 

When evaluating the averaged regret over top 5, we note that when the door is closed, the average regret is fairly similar to first place regret, but when we have open doors, the average regret of the top 5 is lower than the top regret for all models, meaning that it is inherently harder to make robust predictions for the modified problem. However, the neural statistician still sees a benefit over the context-free baselines. Moreover, when we average the top 5, we see that the neural statistician achieves a lower expected regret on the open-door case when compared to the closed door case, but the VAE and baseline models have higher regret. This suggests that there is some benefit to using a pretrained NS is preferable when the image data contains open doors. In practice, such a dataset can be constructed by watching a previous video of that door opening or closing.

\begin{figure}[h]
    \centering
    \includegraphics[scale=0.55]{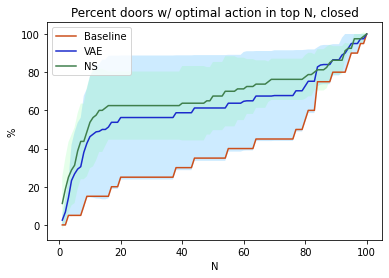}
    \includegraphics[scale=0.55]{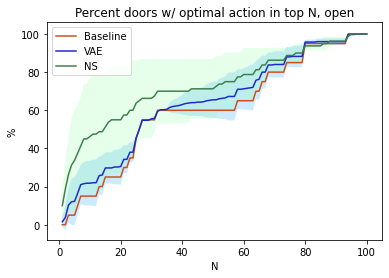}
    \caption{Percentage of doors in which the true optimal action occurs within the top N doors, N going from 1 to 100.}
    \label{fig:hits_breakdown}
\end{figure}

Theoretically, an ideal model would be able to achieve regret close to zero by choosing the action that corresponds with the maximum distance opened in the dataset. However, as can be seen by the results above, all models have difficulty with this task. Figure \ref{fig:hits_breakdown} shows the recall percentage, or the percentage of doors for which the model was able to correctly identify the optimal action as the top action. We see that on average, the neural statistician has overall higher recall for both closed and open door tasks, but the VAE has a much higher variance over different trials than the neural statistician and is still likely to achieve higher recall on the closed-door task. On the other hand, with the open-door task, the NS can identify the top door with over 50\% accuracy within the top 20 actions, which the context-free models are unable to achieve. This is in line with our previous results about the regret values themselves.

% However, the NS achieves lower regret than VAE in the open-door task, indicating that having access to multiple configurations of the door for inference aids in better policy selection. Interestingly, although the open-door task is inherently more difficult, as is seen from the baseline achieving lower regret on the closed-door task, both NS and VAE perform better on the open-door task than on the closed-door task, with the NS on open-door task achieving the lowest regret out of all experimental setups. This is because both the NS and VAE were trained on doors open to all possible configurations. Additionally, this means that a pretrained NS is preferable when the image data contains open doors, so the robot can learn an optimal door-opening policy by watching a previous video of that door opening or closing.

\section{Concluding Remarks}
Both the neural statistician and variational autoencoder models were able to generate faithful reconstructions and realistic samplings when trained on enough data for enough epochs. This finding is not surprising, as having a large enough training set enables the model to learn the underlying probability distribution easily. In fact, it is worth noting that both the NS and VAE are both fairly data-hungry models, a fact that will need to be considered in application. However, the neural statistician provides us a way of separating the latent space into context-level and sample-level parameters, which is useful for interpretability and sampling within a given context.

Finetuning the neural statistician encoder provided a significant performance advantage over the VAE for flipped accuracy, size, handle offset, and angle prediction. The reason for this is likely because learning the context-level parameters of the door enables the neural statistician to correct for discrepancies introduced by the angle of the door. On the other hand, the NS observed worse performance for origin prediction, and the reason for this is likely because the NS was unable to learn that the door location is a context level parameter. In summary, the biggest advantage provided by the neural statistician over the variational autoencoder during parameter inference is that it can draw upon context-level information (from pretraining) to provide a more complete representation of an image that doesn't contain sufficient information on its own (for example, an image of a door open at 90 degrees does not reveal sufficient information about its size).

A pretrained neural statistician shows advantage over a pretrained VAE or a vanilla CNN in reward prediction tasks. %when using image data with a variety of door configurations. Both pretrained models achieved lower regret when the training data consisted of doors at different angle than when the dataset consisted of all closed doors, and both pretrained models consistently achieved lower regret than baseline regardless of finetuning dataset.
This shows that pretraining using a structured latent variable model is advantageous for reward prediction on the doors dataset with limited data. In this investigation, we simulated policy exploration by randomly interacting 100 times with each door, but an application of the neural statistician to more intelligent policy-learning algorithms could potentially show similar trends. Moreover, strategic policy selection methods may yield training datasets with a higher percentage of successful manipulations, allowing the models to achieve lower regrets than in this paper (where the regret is relatively high, but demonstrates a clear trend when comparing across models).

Now that we have demonstrated that the special structure of the neural statistician proves beneficial in learning low-dimensional parameterizations, which in turn provide an advantage in door detection and reward prediction tasks, it is worth investigating the performance of the Neural Statistician model in more generalized datasets. For example, these findings can be applied to more realistic door datasets, datasets where the viewpoint varies slightly, as well as to generalized articulated object datasets (not limited to doors). It was noted previously that the NS requires a large pretraining set to most effectively learn latent variables, but it may not always be feasible to attain such a large pretraining set. Therefore, it is worth investigating how well the NS generalizes to distributional shifts between pretraining and finetuning, such as in an example where the model is pretrained in simulation and finetuned on realistic doors from the application domain.

% \bibliographystyle{unsrtnat}
% \bibliography{sample}

\end{document}